\documentclass[a4paper, 10pt, conference]{ieeeconf}

\IEEEoverridecommandlockouts                       

\usepackage{graphics} 
\usepackage{epsfig} 
\usepackage{mathptmx} 
\usepackage{times} 
\usepackage{amsmath} 
\usepackage{amssymb}  

\usepackage{color}

\usepackage{fullpage}
\usepackage[pdftex]{hyperref}

\usepackage[hyphenbreaks]{breakurl}

\title{\LARGE \bf
3D Vision Guided Robotic Charging Station for Electric and Plug-in Hybrid Vehicles}

\author{Justinas Mi\v{s}eikis$^{1}$,
Matthias R{\"u}ther$^{2}$,
Bernhard Walzel$^{3}$,
Mario Hirz$^{3}$ and 
Helmut Brunner$^{3}$
\thanks{$^{1}$Justinas Mi\v{s}eikis is with Department of Informatics,
University of Oslo, Oslo, Norway{\tt\small justinm@ifi.uio.no}}%
\thanks{$^{2}$Matthias R{\"u}ther is with Graz University of Technology, Institute for Computer Graphics and Vision, Graz, Austria {\tt\small 
ruether@icg.tugraz.at}}%
\thanks{$^{3}$Bernhard Walzel, Mario Hirz and Helmut Brunner are with Graz University of Technology, Institute for Automotive Engineering, Graz, Austria {\tt\small 
\{bernhard.walzel, mario.hirz, helmut.brunner\}@icg.tugraz.at}}%
}

\begin{document}

\maketitle
\thispagestyle{empty}
\pagestyle{empty}

\begin{abstract}

Electric vehicles (EVs) and plug-in hybrid vehicles (PHEVs) are rapidly gaining popularity on our roads. Besides a comparatively high purchasing price, the main two problems limiting their use are the short driving range and inconvenient charging process. In this paper we address the following by presenting an automatic robot-based charging station with 3D vision guidance for plugging and unplugging the charger. First of all, the whole system concept consisting of a 3D vision system, an UR10 robot and a charging station is presented. Then we show the shape-based matching methods used to successfully identify and get the exact pose of the charging port. The same approach is used to calibrate the camera-robot system by using just known structure of the connector plug and no additional markers. Finally, a three-step robot motion planning procedure for plug-in is presented and functionality is demonstrated in a series of successful experiments. 

\end{abstract}

\section{INTRODUCTION}

Nowadays it is common to see electric vehicles and plug-in hybrids on our roads. Worldwide plug-in vehicle sales in 2016 were $773 600$ units, $42\%$ higher compared to 2015~\cite{EVVolume65:online}. For example Norway plans to rule out sales of any combustion engine cars by 2025~\cite{Norwayto43:online}. However, a new problem being faced by EV and PHEV drivers is having an accessible, fast and convenient battery charging, especially when traveling longer distances. It becomes a common problem of fast chargers being occupied even when the car has finished charging, but the owner has not come back yet. For example, Tesla has added an additional idle fee to discourage drivers leaving their cars at the chargers for longer than necessary~\cite{Teslaown67:online}. A solution to avoid this problem and to enable a comfortable fast charging, would be an automated robot-based charging system combined with automated car parking.

\subsection{Charging Ports and Cables}

Worldwide, there are many types of EV and PHEV charging ports, as well as different charging port placement locations on the vehicle. Each one of them has benefits and detriments, and car manufacturers have not decided on a common standard yet. This introduces an additional inconvenience of finding the correct type of charger, or having to carry a number of bulky adapters. 
As long as there is no standard, it would be more convenient to let the charging station detect the correct port type and adapt accordingly.


Another issue is the current weight and stiffness of a quick charging cable. For example, the weight of a CCS-Type 2 charging cable rated for the power up to 200 kW is 2.26 kg/m and outer diameter of 32 mm. With longer cable lengths, this becomes difficult for people to handle, but would not be an issue for a robot~\cite{PHOENIXC32:online}. Cooled charging cables can help to solve this problem by increasing the cable diameter, but these are not yet standard~\cite{IEEEexample:Phoenixcontact_2}.

\subsection{Existing Automated EV Charging Methods}

Automatic charging solutions have been researched both in academic and industrial environments. Volkswagen have presented an e-smartConnect system, where a Kuka LBR-iiwa robot automatically plugs in the vehicle after it autonomously parks in a specific target area (allowing for less than 20 cm by 20 cm error). It is also limited to one charging port type~\cite{esmartCo57:online}.

Tesla have demonstrated a concept of a snake-like robot automatically plugging in their EV, however, no technical details on the charging port localisation or robot operation were revealed~\cite{TeslaUnv99:online}.

The Dortmund Technical University have presented a prototype of the automatic charging system called ALanE. It is based on a robot arm capable of automatically plugging and unplugging a standard energy supply to an electric vehicle. The system is controlled via smartphone. However, full capabilities and flexibility of this concept system are not clear~\cite{Ladesyst44:online}.

\begin{figure}[h]
\centering
\includegraphics[width=0.48\textwidth]{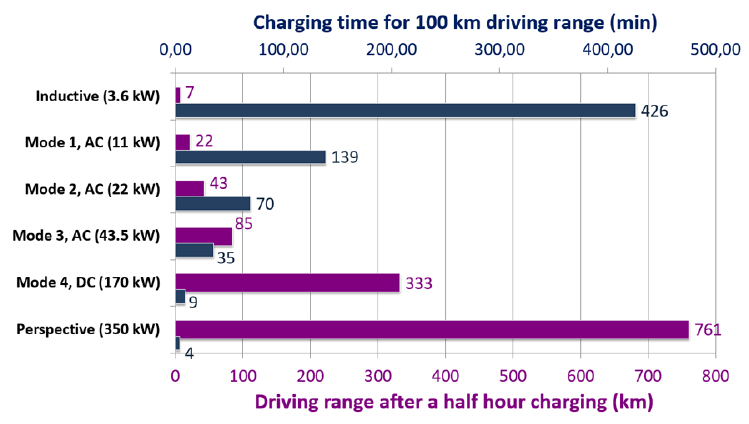}
\caption{Driving distance and charging time comparison of different charging systems~\cite{walzel2016charging}.}
\label{fig:charging_times}
\vspace{-0.2cm}
\end{figure}

The NRG-X concept presents itself as a fully automatic charging solution. It can be adapted to any EV or PHEV and is capable of fast charging. Furthermore, it has a tolerance for inaccurate parking positions. The NRG-X system is based on combination of conductive and inductive charging on the under-body of the vehicle, thus an adapter for the vehicle is necessary. Furthermore, the current concept configuration the charging power is limited to 22 kW~\cite{NRGXAuto61:online}, which results in over 7 times longer charging compared to 170 kW charging~\cite{walzel2016charging} and perspective 350kW~\cite{IEEEexample:CharINe}. Comparisons of the time taken to charge a vehicle using different charging systems is shown in Figure~\ref{fig:charging_times}.

\subsection{Related Research}

Automated charging has been well researched, especially for mobile robots. Typically, there is a custom made charging station, which is localized by the robot either using a direct communication or using computer vision based methods. These methods are normally based on having special markers on the charging station, which are localised in order for the robot to correctly align itself and approach the station. Removing markers would impede the operation~\cite{kartoun2006vision}~\cite{silverman2002staying}~\cite{silverman2003staying}~\cite{luo2005vision}. 

Another concept developed specifically for the detection of charging ports on EVs was based on adding an array of RFID tags on the car. Reading RFID signals allows to find the exact position and orientation of the charging port and plug it in automatically~\cite{oh2015rfid}. However, this still requires modification to the vehicle and would not support non-adapted cars.

\begin{figure}[h]
\centering
\includegraphics[width=0.35\textwidth]{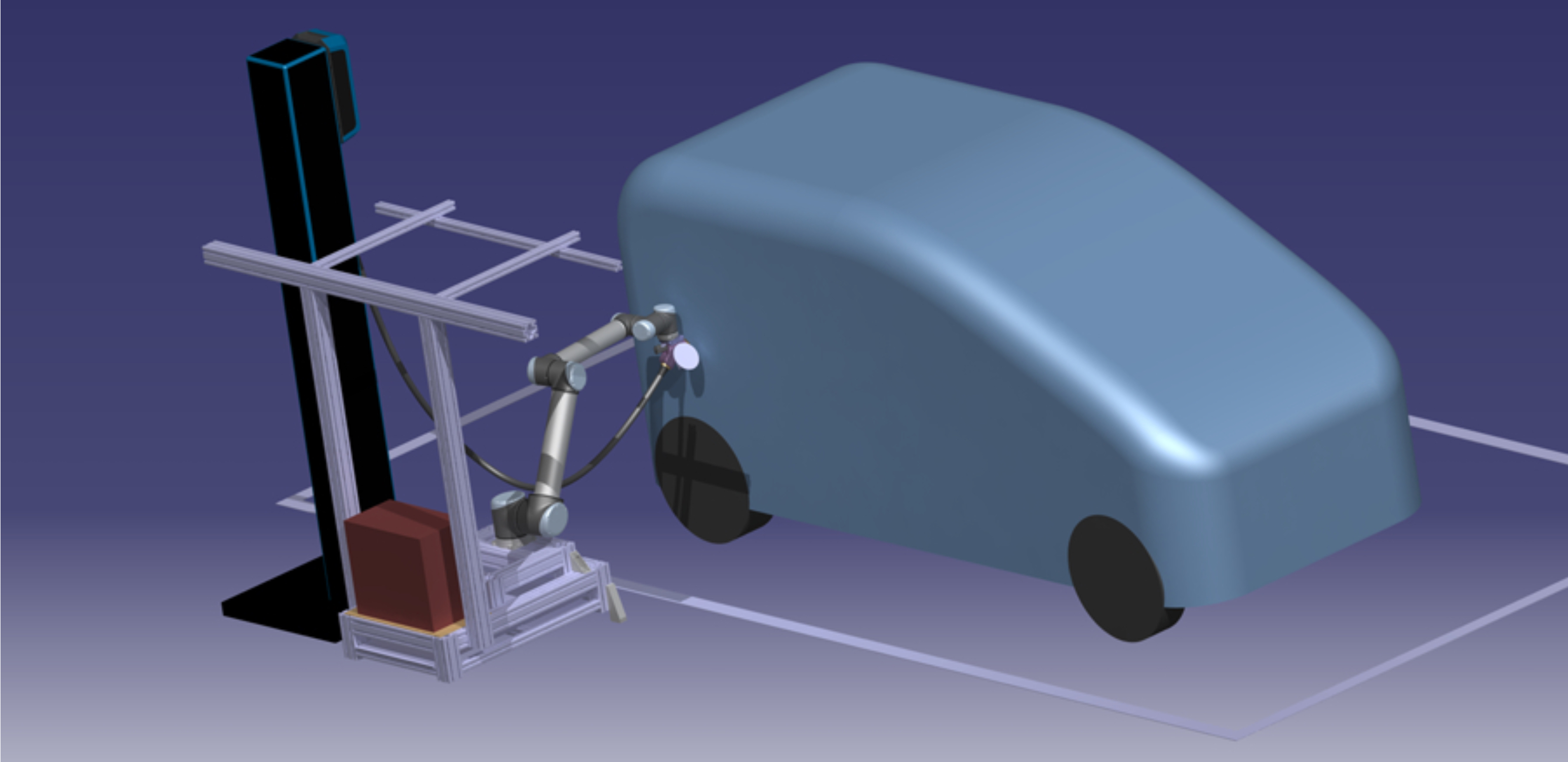}
\caption{CAD model of the robotic charging station concept.}
\label{fig:charging_station}
\vspace{-0.4cm}
\end{figure}

\subsection{Our Method}

We present a conductive robot-based automated charging method for EVs and PHEVs, which does not require any modifications to existing vehicles. First of all, we present a quick eye-to-hand calibration procedure to calibrate the vision sensor and the robot to work in the same coordinate system. It estimates both, the placement of the vision sensor in relation to the robot base as well as between the end-effector and the plug. Then we use shape-based matching and triangulation to locate and identify the charging port of the car and guide the robot, holding a charging cable, to precisely plug in the charger. Once the car is fully charged, the robot will automatically unplug from the vehicle, which will be ready to be driven away. The visualisation of the concept robotic charging station is shown in Figure~\ref{fig:charging_station}.

This paper is organized as follows. We explain the proposed method in Section~\ref{sec:method}. Then we provide our test setup, experiments and results in Section~\ref{sec:experiments}, followed by conclusions and future work in Section~\ref{sec:conclusions}.

\section{METHOD}
\label{sec:method}

\subsection{Detection of the Charging Port}

A majority of the car charging ports are manufactured from texture-less black plastic material, making it difficult to obtain good features in the camera image. Similarly, the measurements made using time-of-flight cameras, which use the projection of infrared (IR) light, are noisy and inaccurate due to IR absorption by the material. As an alternative solution, a stereo-camera setup was used as the vision sensor.


\begin{figure}[h]
\centering
\includegraphics[width=0.40\textwidth]{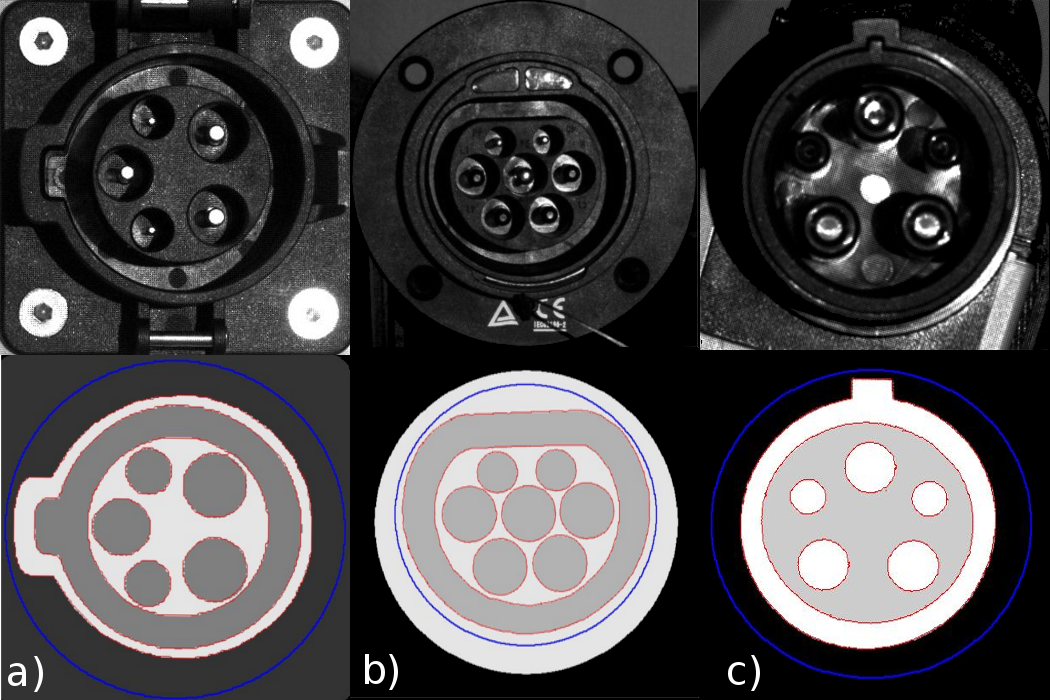}
\caption{Input images, simplified template models and automatically created shape-based templates for matching. Type 2 socket is shown in column a), type 1 socket in b) and type 2 connector plug is shown in c). Blue circles define the area of interest for the model creation and the red outline line defines the created shape model.}
\label{fig:plugs-models}
\vspace{-0.2cm}
\end{figure}

The first step in the detection procedure is to find the location of the charging port in stereo images using shape-based template matching. Models were created for two types of the charging ports as well as the power plug connector, later to be used for eye-to-hand calibration. Figure~\ref{fig:plugs-models} shows the camera images and simplified model images, which are used to automatically generate shape-based templates later to be used for matching. Template matching was performed using a \textit{Halcon Machine Vision} software, which has proven to perform well in given conditions of low-contrast input images~\cite{HALCON–T50:online}. Matching results in a 2D Affine transformation matrix defining the template location in the image.

By taking $x$ and $y$ coordinates of the corresponding object points in images from each of the stereo cameras, the depth information defined by $z$-axis can be calculated. The vision sensor in our setup has both stereo cameras fixed in relation to each other looking slightly inwards, with rotation around $Y$ (vertical) axis. Solving Eq.~\ref{eq:triangulation} provides the real-world coordinates $X$, $Y$ and $Z$ of a point seen by the stereo cameras. Inputs ($x_1$, $y_1$) and ($x_2$, $y_2$) are the point coordinates in camera 1 and camera 2 respectively. Variable $f$ is the focal length of the camera and $b$ defines a baseline (distance) between the stereo cameras. Rotation between the cameras around $Y$-axis is defined by $\theta $.

\begin{equation}
\begin{split}
    &Z_0 = \frac{b}{tan(\theta)} \\
    &Z = \frac{b*f}{x_1-x_2+\frac{f*b}{Z_0}} \\
    &X = \frac{x_1*Z}{f} \\
    &Y = \frac{y_1*Z}{f}
\end{split}
\label{eq:triangulation}
\vspace{-0.2cm}
\end{equation}


After the charging port is found in the input images, stereo triangulation is used to obtain 3D real-world coordinates of the port position, providing 5 to 7 reference points depending on the charging port type. Using the points, a perspective transformation is calculated using the least squares fit method to obtain the exact position and orientation of the charging port in relation to the vision sensor. Least squares fit for finding the orientation optimises for 3 unknowns ($A$, $B$ and $C$), which later are mapped to roll, pitch and yaw angles. The least square error function is defined in Eq.~\ref{eq:least_squares_errorf}, where $x$, $y$ and $z$ are coordinates of the reference points.


\begin{equation}
e(A,B,C) = \sum(Ax + By + C - z)^2
\label{eq:least_squares_errorf}
\end{equation}

Then, the error function is differentiated and set to zero, as shown in Eq.~\ref{eq:least_squares_diff}.

\begin{equation}
\begin{split}
    &\frac{\partial e}{\partial A} = \sum{2(Ax+By+C-z)x} = 0 \\
    &\frac{\partial e}{\partial B} = \sum{2(Ax+By+C-z)y} = 0 \\
    &\frac{\partial e}{\partial C} = \sum{2(Ax+By+C-z)} = 0
\end{split}
\label{eq:least_squares_diff}
\vspace{-0.2cm}
\end{equation}

The resulting linear equations with $3$ unknowns are solved to get the orientation of the object. This can also be seen as 3D plane fitting to the given points. 

\subsection{Marker-less Eye-to-Hand Calibration}

In order to operate the vision sensor and the robot in the same coordinate system, eye-to-hand calibration is necessary. The eye-to-hand calibration estimates the transformation between the vision sensor and the robot base. Using this transformation, the position of any object detected by the vision sensor can be recalculated into the coordinate system of the robot, allowing the robot to move to, or avoid that specific location.

Normally, a well structured object, like a checkerboard of known size and structure is used in the calibration process. However, it requires mounting it on the end-effector of the robot and can still result in additional offsets. We use the known structure of the connector plug and previously presented shape-based template matching with orientation estimation to obtain the precise pose. Eye-to-hand calibration is based on an automatic calibration procedure for 3D camera-robot systems, which uses the calibration method proposed by Tsai et al~\cite{miseikis2016automatic}~\cite{tsai1989new}.

The result of the eye-to-hand calibration are two transformation matrices. The first one defines the the position of the vision sensor in relation to the robot base and the second one defines the position of the end point of the connector plug in relation to the end-effector of the robot.

The marker-less eye-to-hand calibration can be beneficial if the robot is placed on a moving platform, so the relative position between the vision sensor and the robot can change. Furthermore, it would benefit in cases when the robot has interchangeable end-effector attachments with different connector plugs. In both of these cases, recalibration procedure could be done automatically without any reconfiguration.


\subsection{Robot Motion Planning}

Given the limited workspace and all the movements being defined by camera measurements, robot control in Cartesian coordinates was used. The \textit{MoveIt!} framework, containing multiple motion planning algorithms, was used for the  initial testing~\cite{sucan2013moveit}. The best performance in the defined case was demonstrated by the RRT-connect algorithm, which is based on the rapidly exploring random trees ~\cite{kuffner2000rrt}.

In order to get smoother motion execution and more human-like motions, a velocity based controller was used instead of the standard one provided in ROS. Better performance is achieved by calculating and directly sending speed commands to each of the robot joints, thus reducing the execution start time to $50-70$ ms compared to around $170$ ms using the official ROS UR10 drivers~\cite{andersen2015optimizing}.

\subsection{Plugging In Procedure}

\begin{figure}[ht]
\centering
\includegraphics[width=0.44\textwidth]{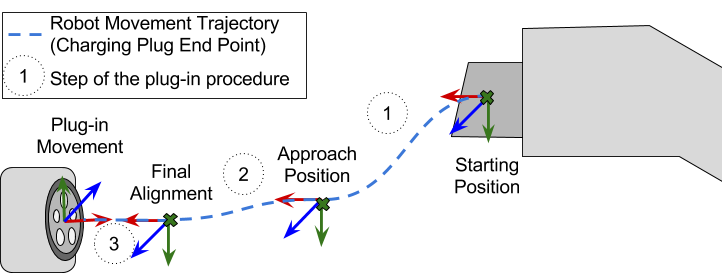}
\caption{Three step plug-in procedure plan. Firstly, the robot moves the connector plug to the \textit{Approach Position}, which lies approximately $0.1$ meter away from the charging port. The second move aligns the Z-axes of the charging port and the plug, and gets the plug just a few millimeters away from the port.
The final plug-in movement
performs the plugging in motion along Z-axis.
}
\label{fig:plug-in_motion_plan}
\vspace{-0.4cm}
\end{figure}

After the pose of the charging port is calculated, the coordinate system is assigned with the origin placed at the center of the plug and Z-axis looking outwards. Similarly, the coordinate system is assigned to the connector plug, which is held by the robot. The goal of the plug-in procedure is to perfectly align connector plug with the charging port, so the last movement is simply along one axis. In order to achieve that, a three-step procedure was used, visualised in Figure~\ref{fig:plug-in_motion_plan}. Firstly, the robot moves the plug at high velocity to the approach position, which is within a $0.1$ meter radius from the charging port. The second step is to reduce the velocity to $10\%$ of the maximum robot joint speed and move to the final alignment position. In this pose, the connector plug and the charging port are fully aligned by their Z-axis and just a few millimeters away from the contact point. The last step is to move at just $2\%$ of the maximum speed along Z-axis and perform the plug-in motion. During this move, the forces and torques exerted on the end effector of the robot are monitored. In case the forces exceed a given threshold, the system is halted to prevent any damage.

\subsection{Unplugging}

After the vehicle is charged fully or to the desired battery level, the robot has to disconnect the charger. Under the assumption that there were no position changes during the charging process, the unplugging procedure was simplified to follow the recorded waypoints of the plug-in procedure in the inverse order. First, the robot gets back to the approach position and then returns to the stand-by position, where it is docked while waiting for the next task. The stand-by position ensures an unobstructed view of the parked vehicle for the vision sensor.

\section{EXPERIMENTS AND RESULTS}
\label{sec:experiments}

\subsection{Experiment Setup}

At the current stage, the testing was limited to the lab environment. The experimental setup consists of an UR10 robot arm, a vision sensor containing stereo cameras and a charging port holder with interchangeable charging ports. Charging port holder has variable height, position and angle to simulate various imperfect parking positions and differences in charging port locations on the vehicle. Two types of the charging ports, Type 1 and 2, have been used, as previously seen in Figure~\ref{fig:plugs-models}.

\begin{figure}[ht]
\centering
\includegraphics[width=0.32\textwidth]{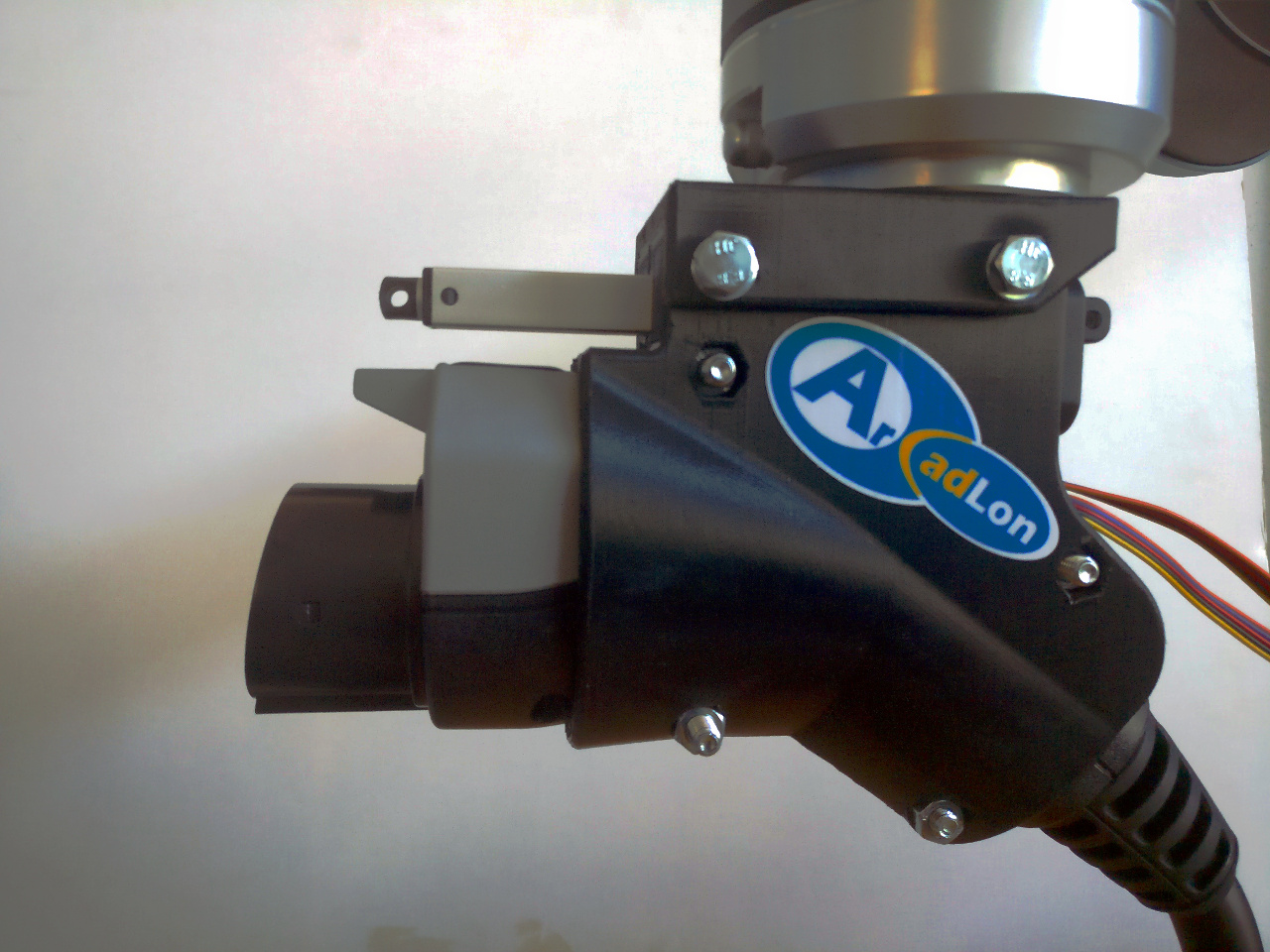}
\caption{Custom 3D printed connector plug holder attached to the end-effector of the UR10 robot.}
\label{fig:ee_attachment}
\vspace{-0.2cm}
\end{figure}

\begin{figure}[ht]
\centering
\includegraphics[width=0.35\textwidth]{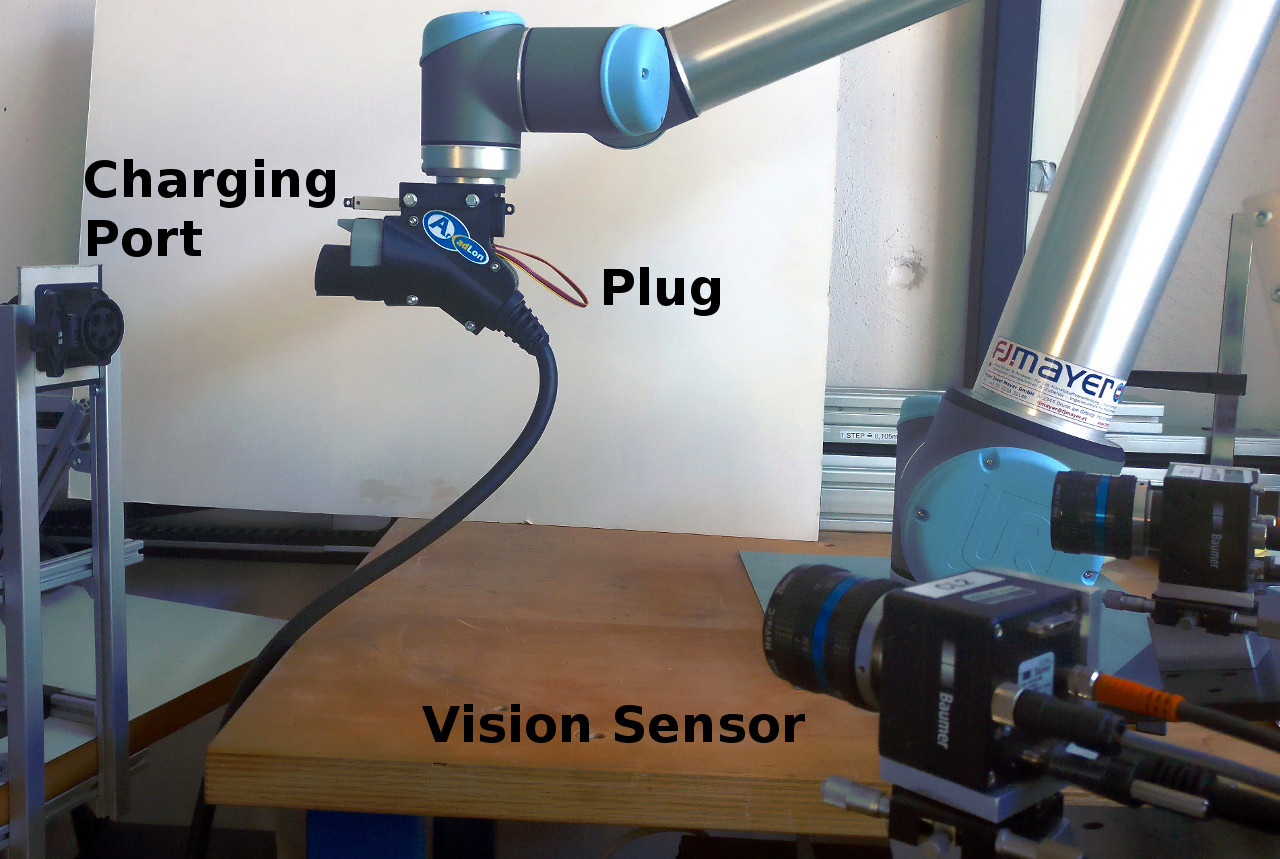}
\caption{The whole experiment setup. On the left the charging port holder can be seen. The robot is holding the connector plug, and the vision sensor made up of two stereo cameras is seen on the right hand side.}
\label{fig:experiment_setup}
\vspace{-0.4cm}
\end{figure}

The connector plug is attached to the end-effector of the robot using a custom 3D printed attachment, shown in Figure~\ref{fig:ee_attachment}. The charging cable is also attached to simulate realistic weight exerted on the robot during the operation. The whole experimental setup is shown in Figure~\ref{fig:experiment_setup}.

The final goal was to locate the charging port using the vision sensor and estimate its pose. Then, the pose is transformed into the coordinate system of the robot and the end point of the connector plug is aligned and plugged in to the charging port. After a brief pause to simulate the charging process, the unplugging movement is performed and the robot moves back to the stand-by position.

Results of each part of the process are discussed separately and followed by the final evaluation of the whole system.

\subsection{Template Matching}

\begin{figure}[h]
\centering
\includegraphics[width=0.48\textwidth]{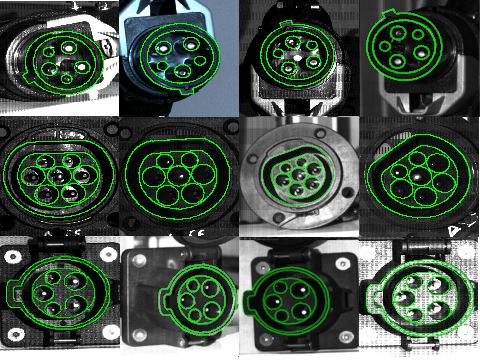}
\caption{Results of the template matching. A high variety of angles and lighting conditions were tested. Viewing angles up to $45^{\circ}$ resulted in successful detection with accuracy dropping beyond that. Row 1: Type 2 connector plug. Row 2: Type 1 socket. Row 3: Type 2 socket.}
\label{fig:halcon_detection}
\vspace{-0.2cm}
\end{figure}

Template matching for Type 1 and Type 2 charging ports as well as the connector plug (Type 2) has worked well for various illumination and angles up to $45^{\circ}$ relative to the viewing angle of the camera. The matching confidence score for good alignment was over $95\%$. The recognition speed on the full camera image was varying between $300 ms$ and $800 ms$. By narrowing down the search area, for example by identifying the darker than average regions in the image, the recognition speed can be reduced to under $150 ms$. The results can be seen in Figure~\ref{fig:halcon_detection}.

The limit for the successful recognition under low illumination or overexposure was when the edges of the socket or plug structure are still visible. The connector plug was made out of more reflective plastic, resulting in a few cases when reflections caused the accuracy issues regarding the rotation. However, these issues were observed very rarely under specific viewing angles, and matching accuracy dropped below $90\%$, so these cases could be easily identified.

\subsection{Eye-to-Hand Calibration}

In the given configuration, the structure of the connector plug was used as a marker for eye-to-hand calibration. During the calibration process it was turned to face the vision sensor, while during the normal operation it faces away from the camera. Furthermore, the outer ring of the plug is angled, so the pins of the plug had to be used as reference points to get the accurate calibration.

\begin{figure}[h]
\centering
\includegraphics[width=0.20\textwidth]{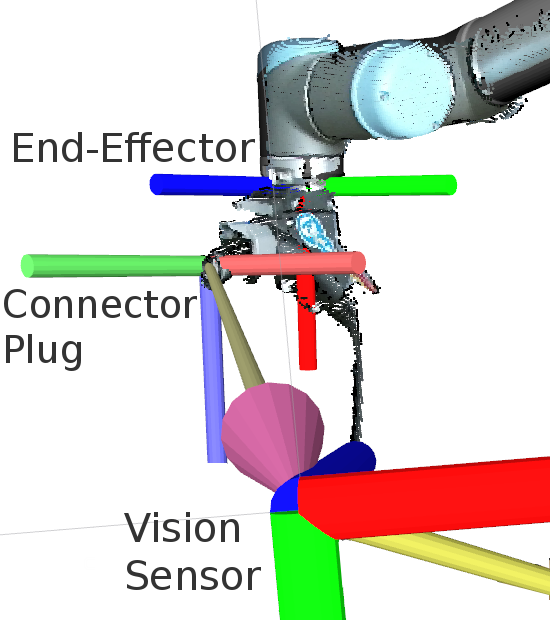}
\caption{Eye-to-hand calibration results. Visualisation of the assigned coordinate frames to the vision sensor, the end-effector of the robot and the end point of the connector plug. Resulting point cloud is overlayed onto the visualisation of the robot model.}
\label{fig:calibTF}
\vspace{-0.4cm}
\end{figure}

The end point of the connector plug was rotated around each of the axis as well as moved to different locations within the field-of-view of the vision sensor. In total, $26$ poses were recorded and used until the calibration converged. Additionally, $3$ instances were discarded because of the incorrect template matching result. The average translation error within the working space was reduced to $1.5mm$, which was sufficient for our application at this stage. Possibly, having more poses would reduce the positional error even further. With the eye-to-hand calibration completed, coordinate frames for the camera position and the end point of the connector plug can be added to the model, as shown in Figure~\ref{fig:calibTF}.

\subsection{Finding Charging Port Pose and Robot Movements}

As the final evaluation, we used the whole process pipeline and analysed whether the plug-in motion was successful or not.

There were $10$ runs executed in total using Type 2 connectors. For the first $5$ runs the charging port was angled at $10^{\circ}$ in relation to the vision sensor, and for the remaining $5$ runs, the angle was increased to $30^{\circ}$.

The robot successfully connected the plug $8$ out of $10$ times. Both failures occurred by missing the rotation of the plug, which were determined by the misalignment of the guidance slot on the charging port. However, the safety stop automatically initialised in both of the cases ensuring that the robot stopped before causing any damage.

\begin{table}[h]
\caption{Summary of the plug-in motion experiments with charging port placed at two different angles}
\label{table:results}
\centering
\begin{tabular}{ |p{0.35cm}||p{3cm}|p{3cm}|}
 \hline
 \textbf{Exp} & \textbf{Charging Port Angle $10^{\circ}$} & \textbf{Charging Port Angle $30^{\circ}$} \\
 \hline
 1 & Success & Success: Misalignment \\ \hline
 2 & Success: Misalignment & \textit{Failed: Missed rotation} \\ \hline
 3 & Success & Success \\ \hline
 4 & \textit{Failed: Missed rotation} & Success: Misalignment \\ \hline
 5 & Success: Misalignment & Success: Misalignment \\ \hline
 \hline
\end{tabular}
\vspace{-0.2cm}
\end{table}

However, even when the plug was successfully inserted in the charging port, there were some alignment issues. In $5$ out of $8$ successful runs, the plug was not fully inserted into the charging port. It was caused by a small angular offset varying between $2^{\circ}$ and $5^{\circ}$. The contact was still made, so the charging process would be successful, however, there was additional strain due to imperfect alignment. The misalignment occurred more frequently during the experiments, where the charging port was placed at  $30^{\circ}$ angle. The results are summarised in Table~\ref{table:results}.

As expected, the unplugging process was successful during all the runs. It simply follows already executed trajectory in the inverse order, meaning that as long as the position of the charging port did not change during the time it was plugged in, there should be no issues with the unplugging process.

\section{CONCLUSIONS AND FUTURE WORK}
\label{sec:conclusions}

We have presented a vision-guided and robot-based automatic EV and PHEV charging station. The goal is to allow automated conductive fast charging of electric and hybrid vehicles and avoid the issue of a charged car taking up the space when it is not necessary.

The presented approach is a combination of multiple methods. First of all, the shape-based template matching is used to identify the charging port type and use the information from stereo cameras to precisely estimate it's position and orientation. The same method is used in the marker-less eye-to-hand calibration, which results in the transformation matrices to be used to convert the position of the charging port from the coordinate system of the vision sensor to the robot. Then, the robot, holding a connector plug, is used to approach and finally plug in the charger cable into the EV or PHEV. Having a precisely estimated orientation is a big challenge and observation of the forces exerted on the end-effector of the robot are necessary to identify any possible misalignment, and stop or readjust if needed. Our approach has proven to work in the lab conditions under indoor illumination and using a custom made charging port holder.

Adding a force sensor to the robot would allow the robot to operate using the impedance controller based on force measurements and adjust it during the plug-in procedure according to the strains observed on the end effector. This would likely to be a solution for the observed cases with misalignment issues.

The project will be continued by improving the connector plug detection accuracy and automating the marker-less calibration procedure, where the robot would perform calibration movements automatically.

Furthermore, current tests were performed under the assumption that the charging port lid or cap was already opened. A linear actuator is already included in the setup, however, it was not used in current experiments. Future work includes finding the charger lid, identifying it's opening mechanism and using the robot to open and close it for the charging process. This would also require identification of the vehicle model to indicate the correct parking position and localise the approximate position of the charging port.

With the test electric vehicle to be delivered in the near future for testing purposes, the system will be evaluated on the real EV in the garage setup and outdoor tests. Communication between the vehicle and the charging station is also under development and this will enable the combination of the robot-based charging system with autonomous parking functions.

\addtolength{\textheight}{-12cm}   








{\small
\bibliographystyle{IEEEtranS}
\bibliography{IEEEexample}
}

\end{document}